%% file: main.tex
\documentclass[runningheads]{llncs}
\usepackage{graphicx}
\usepackage{amsmath,amssymb,amsfonts} %

\usepackage[utf8]{inputenc} %
\usepackage[T1]{fontenc}    %

\usepackage[font=small,skip=4pt]{subcaption}
\usepackage[font=small,skip=4pt,tableposition=top]{caption}

\usepackage[numbers,sort,compress]{natbib} 

\usepackage{booktabs}  
\usepackage{multirow}
\usepackage{microtype} 
\usepackage{nicefrac}  
\usepackage{newtxmath} 

\usepackage{hyperref}

\usepackage[dvipsnames]{xcolor}         
\usepackage{graphicx}
\usepackage{pifont}%
\newcommand{\cmark}{\ding{51}}%
\newcommand{\xmark}{\ding{55}}%
\usepackage[dvipsnames]{xcolor}
\usepackage{listings}

\usepackage{comment}
\usepackage[width=122mm,left=12mm,paperwidth=146mm,height=193mm,top=12mm,paperheight=217mm]{geometry}

\hypersetup{
     colorlinks   = true,
     citecolor    = blue
}

\begin{document}
\mainmatter

\title{Text-Conditioned Resampler For Long Form Video Understanding} 

\titlerunning{\em TCR}
\authorrunning{B.~Korbar et al.}

\author{Bruno Korbar\inst{1,2} \and
Yongqin Xian\inst{2} \and
Alessio Tonioni\inst{2} \and
Andrew Zisserman\inst{1, 3}
\and
Federico Tombari\inst{2,4}
}
\institute{Visual Geometry Group, University of Oxford\\ *\email{korbar@robots.ox.ac.uk} \and
Google, Zurich \and
Google DeepMind, London \and
TU Munich, Munich
\\
}

\maketitle

\begin{abstract}
  In this paper we present a text-conditioned video resampler (TCR) module that uses a pre-trained and frozen visual encoder and large language model (LLM) to process long video sequences for a task. TCR localises relevant visual features from the video given a text condition and provides them to a LLM to generate a text response.  Due to its lightweight design and use of cross-attention, TCR can process more than 100 frames at a time with plain attention and without optimised implementations. We make the following contributions: (i) we design a transformer-based sampling architecture that can process long videos conditioned on a task, together with a training method that enables it to bridge pre-trained visual and language models; (ii) we identify tasks that could benefit from longer video perception; and (iii) we empirically validate its efficacy on a wide variety of evaluation tasks including NextQA, EgoSchema, and the EGO4D-LTA challenge.
\end{abstract}

\input{_body}

\clearpage

\setcounter{page}{1}
\appendix
\chapter*{Appendix}
\input{_supp}

\bibliographystyle{abbrvnat}
{\footnotesize
\bibliography{main, vgg_local}}

\end{document}

%% file: _body.tex

\section{Introduction}
\label{sec:intro}

The development of visual-language models (VLMs) advanced exponentially in the past few years: new models pre-trained with increasingly larger scale, in terms of the number of parameters and size of the training set, continue pushing forward the state of the art on multiple tasks every couple of months. These models often have the ability to reason about the relationships of objects in their environment through natural language, often in an interactive fashion. This capability is appealing for multiple video applications. For example, it would be helpful for a model to be able to answer questions about a video: ``Does this recipe use eggs?'', ``what does he do after he removes the tire?'', etc. It is also appealing for users of augmented-reality devices: for example to be able to answer ``where did I leave my phone?''. Unfortunately, the computational requirements of such models made them impractical for use in video applications as the memory requirement rises quadratically with the input size. Furthermore, to our knowledge, a large-enough source of even loosely labelled video data for training such a model from scratch does not readily exist. 

\begin{figure}[t]
    \centering
    \includegraphics[width=\textwidth]{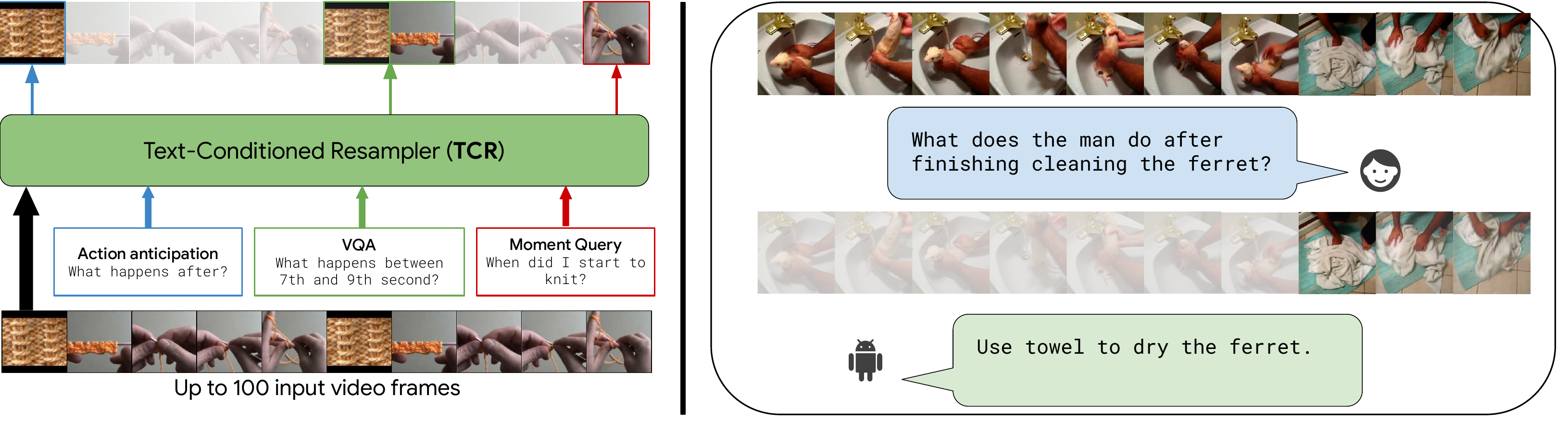}
    \vspace{-15pt}
    \caption{TCR resamples visual features that are relevant for the downstream tasks before passing them to the LLM. A qualitative example can be seen on the right.}
    \label{fig:splash}
    \vspace{-20pt}
\end{figure}

That is why we are specifically interested in a subset of these models that are not trained from scratch, but rather `bridge' pre-trained models via different types of `visual-to-language adapter modules'~\cite{li_blip-2_2023, alayrac_flamingo_2022, sevilla2021only}. The advantages of this approach, as opposed to training the model from scratch, are numerous: Only a small number of parameters are trained, which makes the memory footprint smaller; it allows us to utilise the capabilities of large visual backbones without overfitting to the downstream task; as well as to leverage the vast amount of knowledge stored in the LLM without suffering common limitations of smaller-scale fine-tuning such as catastrophic forgetting. 
Only a few of these models are trained on videos~\cite{alayrac_flamingo_2022, kuo_mammut_2023, yu2023sevilla}, and these can usually ingest only a small number of frames -- typically anywhere between 4 to 32. Allowing a large number of video frames to interact with text is demonstrably beneficial~\cite{sevilla2021only, mangalam2023egoschema} in visual models, thus, a relatively simple way of increasing the model performance is to increase the number of frames the model sees.

In this paper we present a {\em Text-Conditioned Resampler} (TCR), an architecture and pre-training method that tackles all of the challenges mentioned above: it is a reasonably lightweight, low-dimensional adapter which acts as an information bottleneck between visual and language models. As shown in Figure~\ref{fig:splash} (left), it is able to process over a 100 (and up to 180) frames at a time, selecting the most relevant frame features to pass to the LLM based on the ``conditioning'' text. TCR allows us to focus on analysing videos with longer temporal span, and identify gains that could be made on longer videos. Right side of Figure~\ref{fig:splash} illustrates an application of our model. This new method allows us to analyse aspects of video datasets we've never been able to before. Specifically, we were able to look at how many frames it takes for a VLM to solve a task, and to determine if increasing the temporal span of perceived video actually brings benefits in terms of performance. We found that increasing temporal span does improve results on the moment-queries EGO4D challenge, and allows us to set the state-of-the-art (SOTA) on long-video question answering on the validation sets of EgoSchema dataset~\cite{mangalam2023egoschema} and EGO4D long-term forecasting challenges, as well as on the temporally-sensitive NextQA dataset~\cite{xiao2021next}.

\begin{figure}[h]
    \centering
    \includegraphics[width=0.9\textwidth]{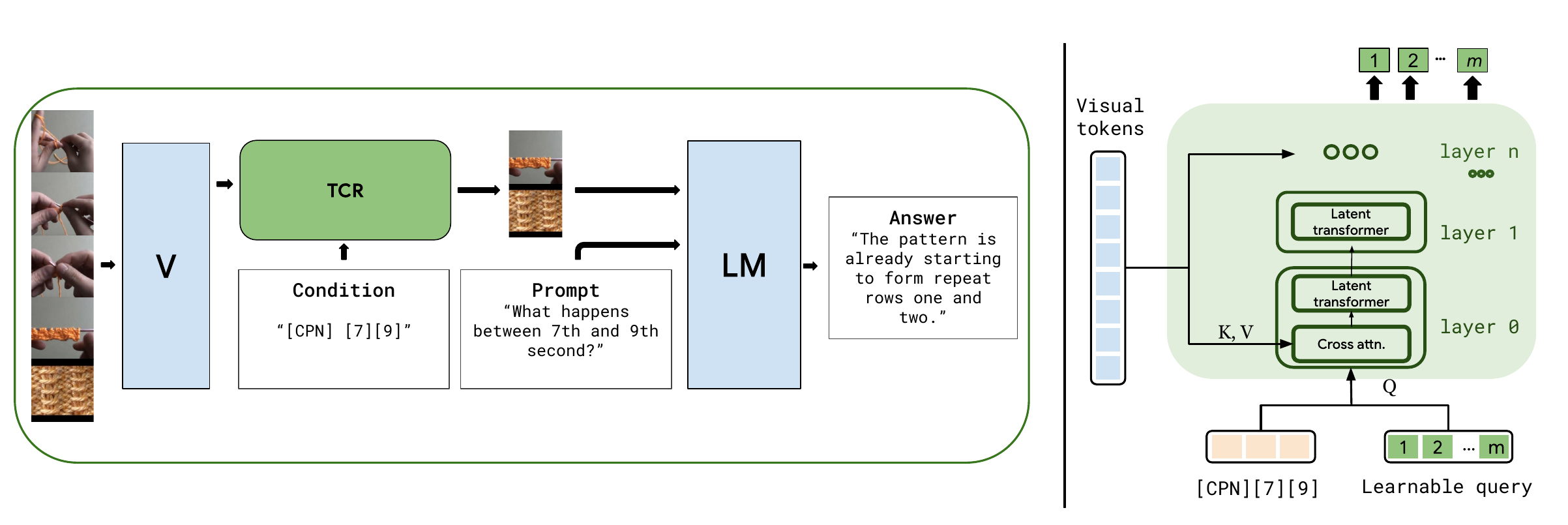}
    \vspace{-5pt}
    \caption{Left: overview of how TCR integrates in a VLM in order to process long videos. A long (30-120 frames) sequence from a visual encoder (V) is resampled to a fixed-length sequence fed to a language model. \texttt{[CPN]} indicates special token for captioning; \texttt{[7][9]} is a representation of tokenised time steps. Right: details of the TCR module. Elements in blue are kept frozen. Best viewed in colour.}
    \vspace{-5pt}
    \label{fig:model}
\end{figure}


\section{Text-Conditioned Resampler (TCR)}
\label{sec:method}
In the following section we describe the model and the training procedures used for training a video-specific VLM able to handle very long video sequences. 


\subsection{Model}
At a high level, the input to the TCR consists of video frames processed by a visual encoder and embedded text tokens. It outputs a fixed-length sequence of embeddings that, together with a text prompt, is consumed by a language model. The text specifies (conditions) the task, and the TCR selects different visual features according to the task and transforms them to be suitable for input to the language model. Finally, the language model generates the text response to the specified task. Architecture overview is given in Figure~\ref{fig:model} on the left.


\noindent \textbf{Overview: }  The visual inputs consist of RGB frames of the video that are ingested by a pre-trained frozen ViT-g~\cite{EVA-CLIP} model to obtain visual embeddings. Temporal encodings are added to them. The conditioning text tokens are prefixed with a learnable special token specifying the task the model is trying to solve and concatenated with a set of learnable query vectors. The queries and text interact with each other through self-attention layers, and interact with the frozen visual features through cross-attention layers (inserted every other transformer block). Output query vectors are then concatenated with an optional text prompt, and passed through a frozen Flan-T5 language model~\cite{chung2022scaling}. The TCR module is illustrated in Figure~\ref{fig:model} on the right. We treat it as a plug-in replacement for the Q-former in the BLIP2~\cite{li_blip-2_2023} architecture to enable handling of very long frame sequences as input.

The key design choices are: (i) the interaction of the query sequence with the visual features is only through cross-attention. This enables the TCR to ingest very long sequences (as it is not limited by the quadratic complexity of vanilla self-attention); and (ii) the output is a fixed length set (the transformed query vectors), so that the input to the language model is only a small number of tokens, irrespective of the length of the video sequence. Following these design principles we are able to significantly reduce the number of input tokens that the LLM needs to process with obvious gains in terms of inference time and memory requirements compared to full self-attention over all frame tokens.


\noindent \textbf{How does the TCR differ from the Flamingo Resampler and Q-former?}
These design decisions build on the architectural innovations of
the Perceiver resampler in Flamingo~\cite{alayrac_flamingo_2022} and the Q-former in BLIP-2~\cite{li_blip-2_2023}.
However, there are a number of differences:
(i) While Q-former is trained on images, TCR is optimised for video from the ground up -- all training stages are done on videos. This is important as the TCR must learn to sample visual features from video frames conditioned on the task. (ii) TCR uses lower dimensional features than either Q-former or Perceiver Resampler (512 vs 768 vs 1536) and an overall smaller number of parameters (69M vs 188M). This is important as it allows us to process far longer video sequences. (iii)
While TCR cross-attends visual features to text embeddings and learnable queries, Perceiver Resampler concatenates visual-embeddings and queries in a key-value pair, which makes the computation more expensive as it computes cross-attention and self-attention in a single pass. We keep the operations separate (i.e.\ first cross-attending text-query sequence with the video, and then self-attending the text-query sequence). This reduces per-layer computational requirements allowing us to increase video sequence length. These differences lead to a novel capability of processing many more frames at once, which subsequently leads to superior performance on downstream tasks.


\noindent\textbf{Conditioning sequence construction: }
Most tasks can be represented as a basic  Question and Answer (QA) pair. Inspired by multi-task language models~\cite{raffel2020exploring}, we adopt a generic \texttt{ [ST][task prompt][learnable query] } input structure  (where \texttt{ [ST] } is a task-specific special token, \texttt{ [task prompt] } is, for example, a question in a QA, and \texttt{ [learnable queries] } are passed on to the LLM). We prefix a special task token (\texttt{[CPN]}, \texttt{[TRG]}, \texttt{[QA]}, \texttt{[STG]}) for captioning, temporal grounding, question-answering, and spatio-temporal grounding respectively) to the task prompt, depending on what task the model is solving. Figure~\ref{fig:model} shows an example wih the \texttt{[CPN]} task-specific special token.
Since, in principle, {\em all} tasks can be formulated as QA (and would be specified in the model as \texttt{[QA][question text]}), why are special tokens used? We found that using tokens improves overall performance while making the model easier to train and reducing the sequence length required for conditioning the sampler (as opposed to spelling out the task in text).


\subsection{Training}
\label{sec:training}
Recent works have shown that contrastive learning yields visual representations for video frames that perform better in discriminative tasks than training in a purely generative fashion~\cite{Yu_VideoBLIP, kuo_mammut_2023}. Training models with a generative loss, however, seems to be crucial for developing reasoning regarding temporal grounding of unconstrained videos as well as the semantic relationship between text structure and video~\cite{yang2023vid2seq, alayrac_flamingo_2022}. Hence, we separate our training in three distinct stages: (i) initialisation, where we train TCR without the LLM; (ii) pre-training, where we train TCR in conjunction with the LLM; and later, (iii) a task-specific fine-tuning. Note that the only thing we're training is the TCR module -- the visual encoder and LLM remain frozen throughout. Initialisation and pre-training stages are done on the YTT-1B dataset~\cite{yang2023vid2seq}. Videos in this dataset are annotated by the transcribed speech sentences and their corresponding timestamps that are either user-generated or automatically generated via automatic-speech recognition. Speech in such videos is rarely visually grounded~\cite{ko2022videoalign, Han20a}, however, because our model can see the video sequence surrounding the annotated segment, it is well suited to implicitly learn the temporal grounding. We describe training stages below.

\begin{table}[t]
\centering
\caption{\small Effect of initialisation and pre-training stages on NextQA question answering and NLQ task. For NextQA, we use shortened fine-tuning procedure (see Section~\ref{sec:nextqa}) and vary the checkpoints used. For NLQ, we evaluate on TCR w/LLM.}
\label{tbl:abltrain}
\vspace{-5pt}
\begin{tabular}{@{}lccccc@{}}
\toprule
\multirow{2}{*}{Init} & \multicolumn{3}{l}{Pre-training} & \multirow{2}{*}{\begin{tabular}[c]{@{}c@{}}NextQA\\ Acc\end{tabular} $\uparrow$}& \multirow{2}{*}{\begin{tabular}[c]{@{}c@{}}NLQ\\ MR@1\end{tabular}  $\uparrow$} \\
 & (i) & (ii) & (iii) &  \\ \midrule
\cmark & \cmark & \cmark & \cmark & \textbf{66.1} & \textbf{11.42} \\ \midrule
\cmark & \xmark & \xmark & \xmark & 52.1 & 7.88 \\
\xmark & \cmark & \cmark & \cmark & 63.3 & 9.41 \\ \midrule
\cmark & \cmark & \xmark & \xmark & 64.1 & 8.94 \\
\cmark & \cmark & \cmark & \xmark & 65.6 & 9.37 \\
\cmark & \xmark & \cmark & \xmark & 63.4 & 8.91 \\
\cmark & \cmark & \xmark & \cmark & 64.2 & 8.13 \\ \bottomrule
\end{tabular}
\vspace{-5pt}
\end{table}


\noindent \textbf{Initialisation (without LLM): }To initialise our model, we follow BLIP2~\cite{li_blip-2_2023}'s \textit{image-text contrastive} and \textit{image-text matching} objectives. Contrastive objective maximises mutual information between TCR text output, and learnable queries which are cross-attended with a video. Text and learnable queries are passed together to TCR. Their mutual attentions masked in such a way that text only attends to itself, while the learnable queries are cross-attended to the video frames and then self-attended to themselves. We compute the average of text queries to get a text representation $t$, and compare it pairwise with all learnable queries. Query with maximum similarity to $t$ is denoted as $q$. We then align the representations $t$ and $q$ by contrasting each positive pair with in-batch negative pairs. At this stage TCR is \textit{not} text conditioned. 
Image-text matching objective (video-text matching in our case) primes the model for text-conditioning. Both learnable queries and text are passed through TCR together, without attention masking. A binary classifier predicting whether the video and text are matching or not is applied to each of the learnable queries and predictions are averaged to obtain a final matching score. The negatives are sampled in-batch, following~\cite{li_blip-2_2023}.   

We skip the \textit{generative} training step of~\cite{li_blip-2_2023}, as our model is neither designed nor initialised from a language model, and we found no measurable benefit from this training stage. The reader is referred to the original paper~\cite{li_blip-2_2023} for in-depth description of attention-masks and losses used during each of the objectives.


\noindent \textbf{Pre-training (with LLM): } The goal of pre-training is twofold: first, to semantically and temporally align TCR's output with the expected input of the LLM, and second to train TCR's self-attention layer to attend to specific task-specifying special tokens and text conditioning tokens. We do this by training it on three tasks. (i) given an untrimmed video and annotated sentence, we ask it to retrieve \textit{when} the sentence occurred; (ii) given the untrimmed video and a timestep, we ask the model to fully caption that particular segment; (iii) given the untrimmed video and a text sequence corrupted in multiple ways, we ask it to correct the sequence. All tasks are supervised by applying the generative loss on the outputs of the LLM. The examples of these tasks on an example from YTT dataset can be seen in the supplementary material. The effects of these training stages can be seen in Table~\ref{tbl:abltrain}.


\noindent \textbf{Fine-tuning: }
After these two stages, TCR achieves competitive results on downstream tasks while still being a generalist model. However, as our pre-training dataset is comprised mostly of low- and mid-quality videos with noisy automatic annotations, we observe significant improvements through fine-tuning for a specific task. The goal of fine-tuning is to align the TCR with the domain of the downstream task in question. Only the TCR module and its vocabulary are fine-tuned, while the visual encoder and the LLM are kept frozen.
Fine-tuning is performed on each of the downstream datasets and is described in the results section for each dataset, while hyperparameters and ablation of the performance with or without fine-tuning are given in the supplementary.


\subsection{Model details}

\noindent\textbf{Video sequence construction: }We extract visual representations ($14\times14$ patches from frames with $224^2$ resolution) using ViT-g~\cite{EVA-CLIP}, and add temporal embeddings. In order to reduce memory consumption, for every other frame we drop random 50\% of its patches. Recent work~\cite{Han22b,tong2022videomae} has shown no significant loss in performance when random patches have been dropped. 


\noindent\textbf{LLM sequence construction: }We follow BLIP2 in the way we construct the input sequence~\cite{li_blip-2_2023}. We concatenate the output of the TCR module together with a \texttt{<BOS>} (beginning of sentence) token and the instruction context tokens (for example question in VQA, or previous action sequence together with instruction for EGO4D action prediction). 


\noindent\textbf{TCR architecture details: }
TCR is based on a transformer-decoder module~\cite{vaswani2017attention}, consisting of 4 transformer blocks with 8 attention heads and hidden dimension equal to 512. Blocks 0 and 2 contain cross-attention layers. For each task we use $128$ 512-dimensional queries. These choices were tuned based on the downstream performance on NextQA validation set and then kept fixed.


\section{Experiments}
\label{sec:results}

\begin{figure}[t]
\begin{subfigure}{0.5\textwidth}
\includegraphics[height=1.1in]{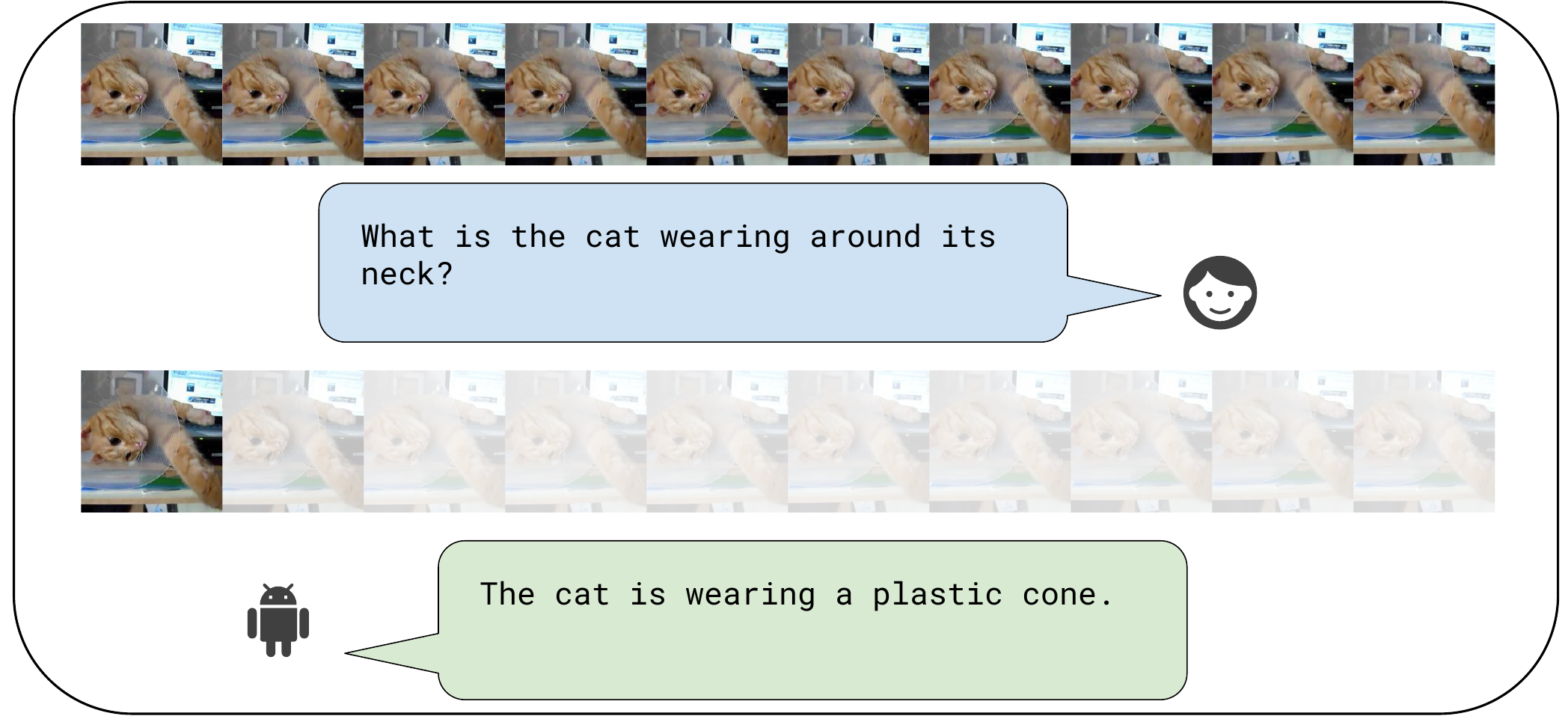}
\end{subfigure}
\begin{subfigure}{0.5\textwidth}
\includegraphics[height=1.1in]{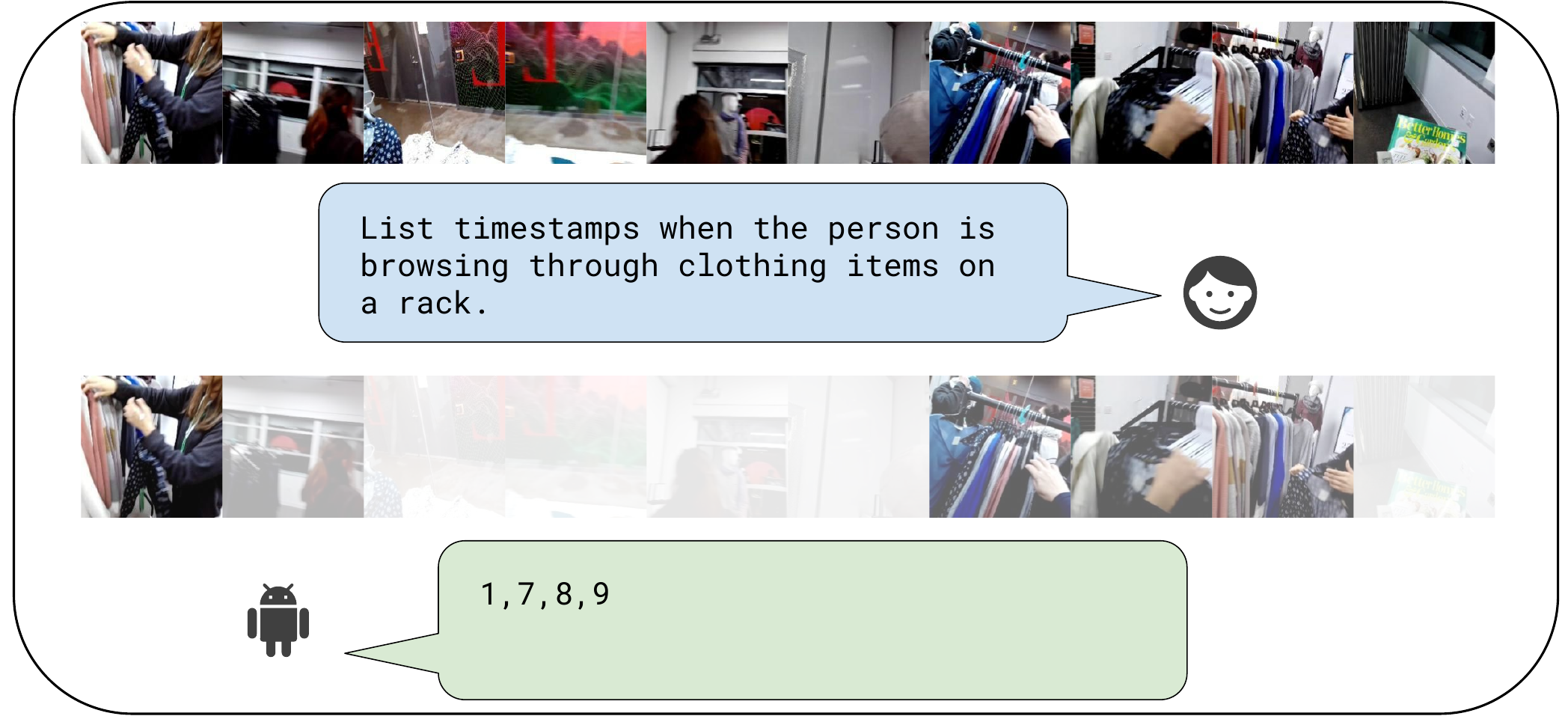}
\end{subfigure}
\begin{subfigure}{0.5\textwidth}
\includegraphics[height=1.1in]{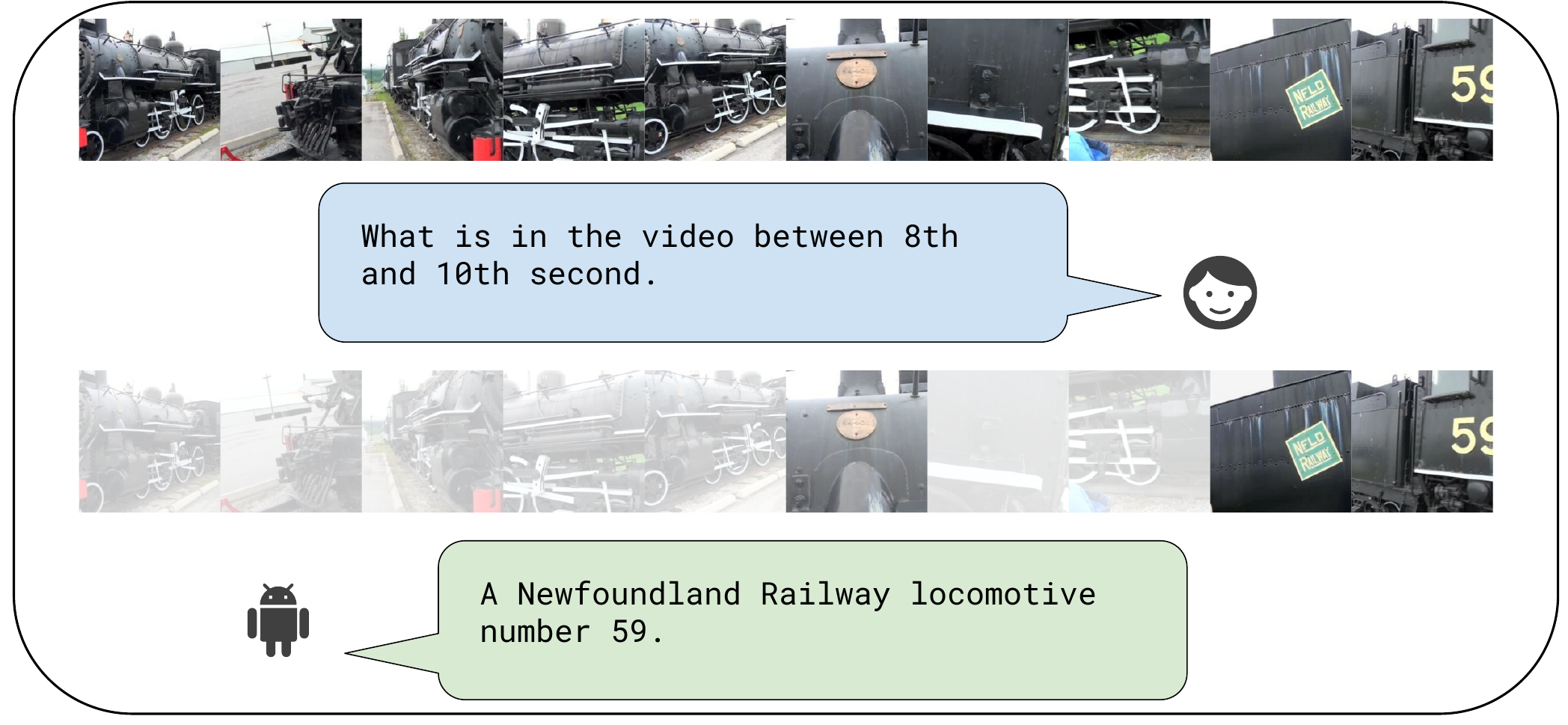}
\end{subfigure}
\begin{subfigure}{0.5\textwidth}
\includegraphics[height=1.1in]{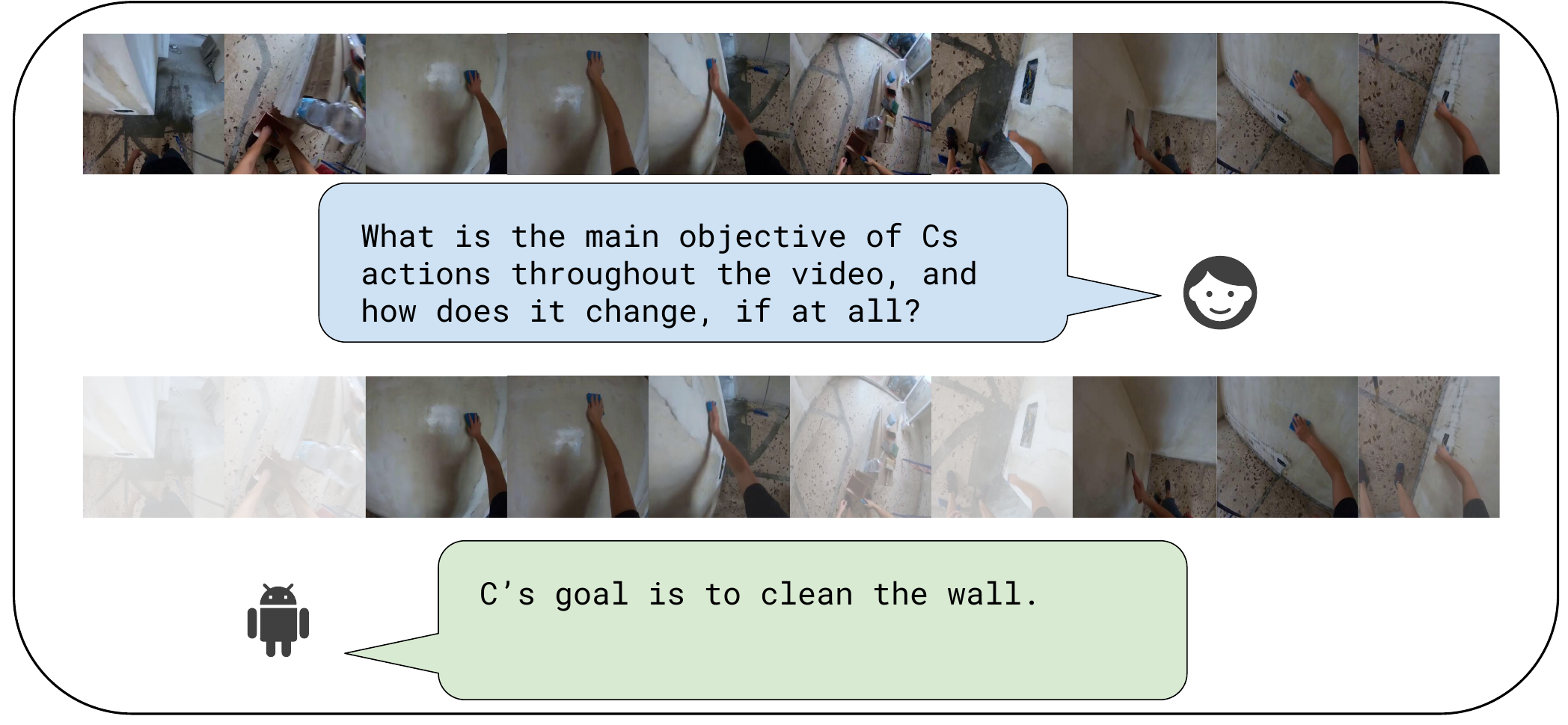}
\end{subfigure}
\caption{\small Examples of our model responding to various textual prompts taken from NextQA, EGO4D-MR, and YTT datasets. The opacity of the images in the second row is correlated to the mean patch attention score for that frame. Note that frames are subsampled and the TCR conditioning is not included for clarity.}
\label{fig:examples}
\end{figure}

 In the following section, we conduct a set of experiments with a baseline VLM and TCR in order to determine which tasks benefit from having access to longer or denser video sequences, and compare the results to the SOTA. Specifically, we analyse the datasets in section~\ref{sec:taskanal}. We compare the results to the state of the art in sections~\ref{sec:nextqa},~\ref{sec:egoschema} and~\ref{sec:ego4d}, and we present ablation of model decisions in section~\ref{sec:ablation}. Qualitative results can be seen in Figure~\ref{fig:examples}.

 
 \noindent \textbf{Datasets: }We evaluate the following datasets: Kinetics400~\cite{kay2017kinetics} containing around 260k 10s videos with human-action labels and Countix, a subset of Kinetics where actions are annotated with the number of repeats (e.g.\ how many time a push up is repeated)~\cite{dwibedi2020counting}. MSR-VTT~\cite{xu2016msr-vtt}, a large scale video captioning dataset. NextQA, a manually annotated video-question-answering dataset where the model is asked to answer questions regarding temporal actions in a multiple-choice fashion~\cite{xiao2021next} from (on average) 44s long videos. Finally, since egocentric videos are a new frontier in effective long-term video understanding, we evaluate on two diverse challenges from EGO4D~\cite{Ego4D2022CVPR}. EGO4D videos are often minutes long, containing both fine-grained actions as well as long-term interactions~\cite{Ego4D2022CVPR}.


\noindent\textbf{Baseline: }
We use a fixed BLIP2~\cite{li_blip-2_2023} VLM as a baseline throughout our experiments. BLIP2 is not trained on videos, but it has been shown that it can be adapted to videos~\cite{Yu_VideoBLIP, damonlpsg2023videollama} and we follow~\cite{Yu_VideoBLIP} to do so. BLIP2 can only process up to 8 frames at a time, so for the task where it can be done, we average predictions over multiple 8-frame video clips extracted at 1fps (noted as `BLIP2(Avg.)'). These aggregation methods, however, can introduce unwanted noise~\cite{sevilla2021only, korbar2019scsampler}.


\noindent \textbf{Modelling longer sequences: } The design of TCR allows a VLM to ``see'' more frames than ever before. Therefore, we also present results using TCR which uses the same visual encoder and LLM as BLIP2 but is able to process \textit{all} the frames at once. With TCR, each video can be processed in a single forward pass thus eliminating the effects of subsampling or averaging.


\subsection{Video task analysis}
\label{sec:taskanal}
\begin{figure}[t]
    \centering
    \subcaptionbox{ \label{sfig:nextqa}}[.3\textwidth]{\includegraphics[height=1in, trim=0 0 0 10,clip]{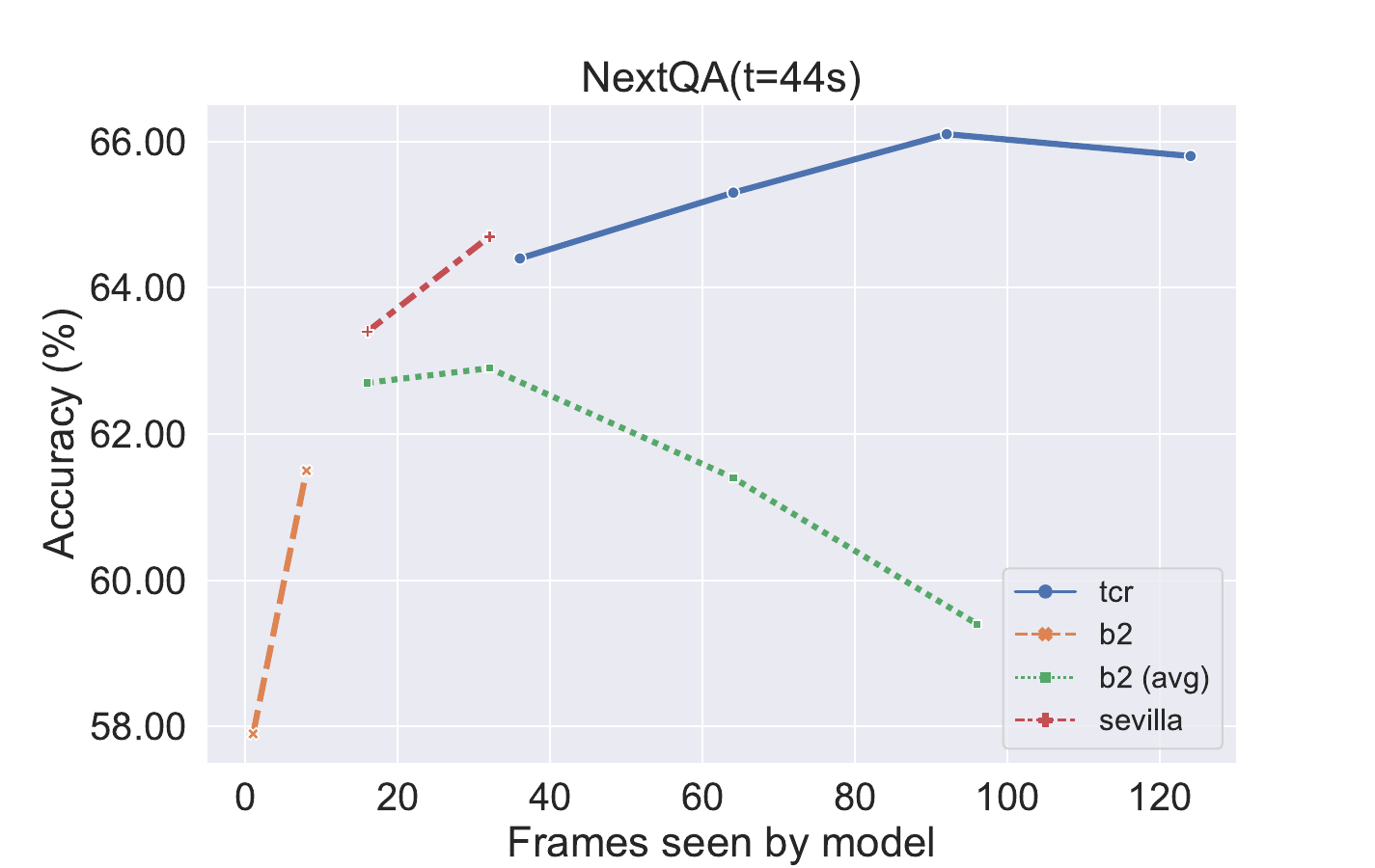}}
    \subcaptionbox{\label{sfig:egoschema}}[.3\textwidth]{\includegraphics[height=1in, trim=0 0 0 0,clip]{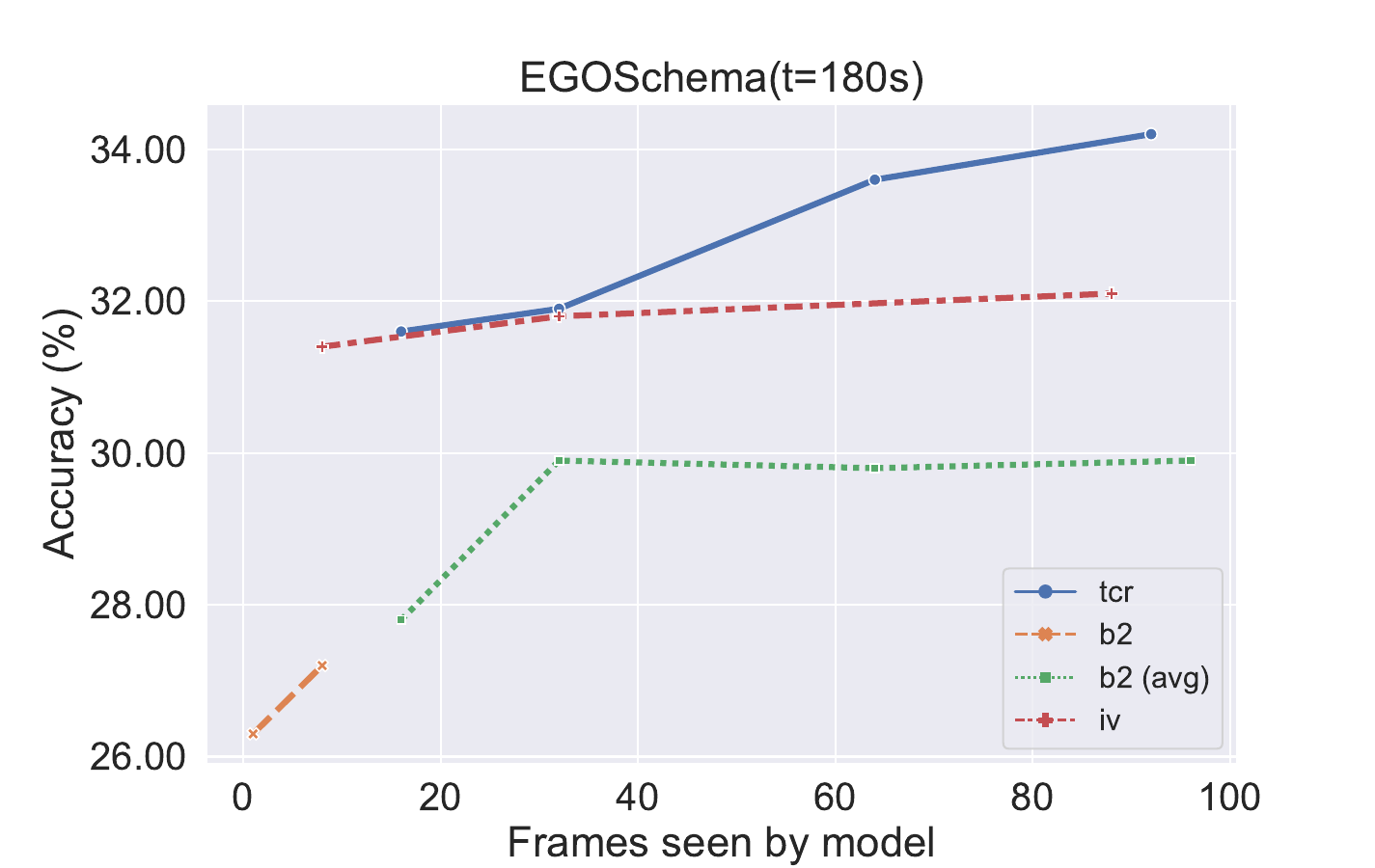}}
    \subcaptionbox{ \label{sfig:ar}}[.3\textwidth]{\includegraphics[height=1in, trim=0 0 0 0,clip]{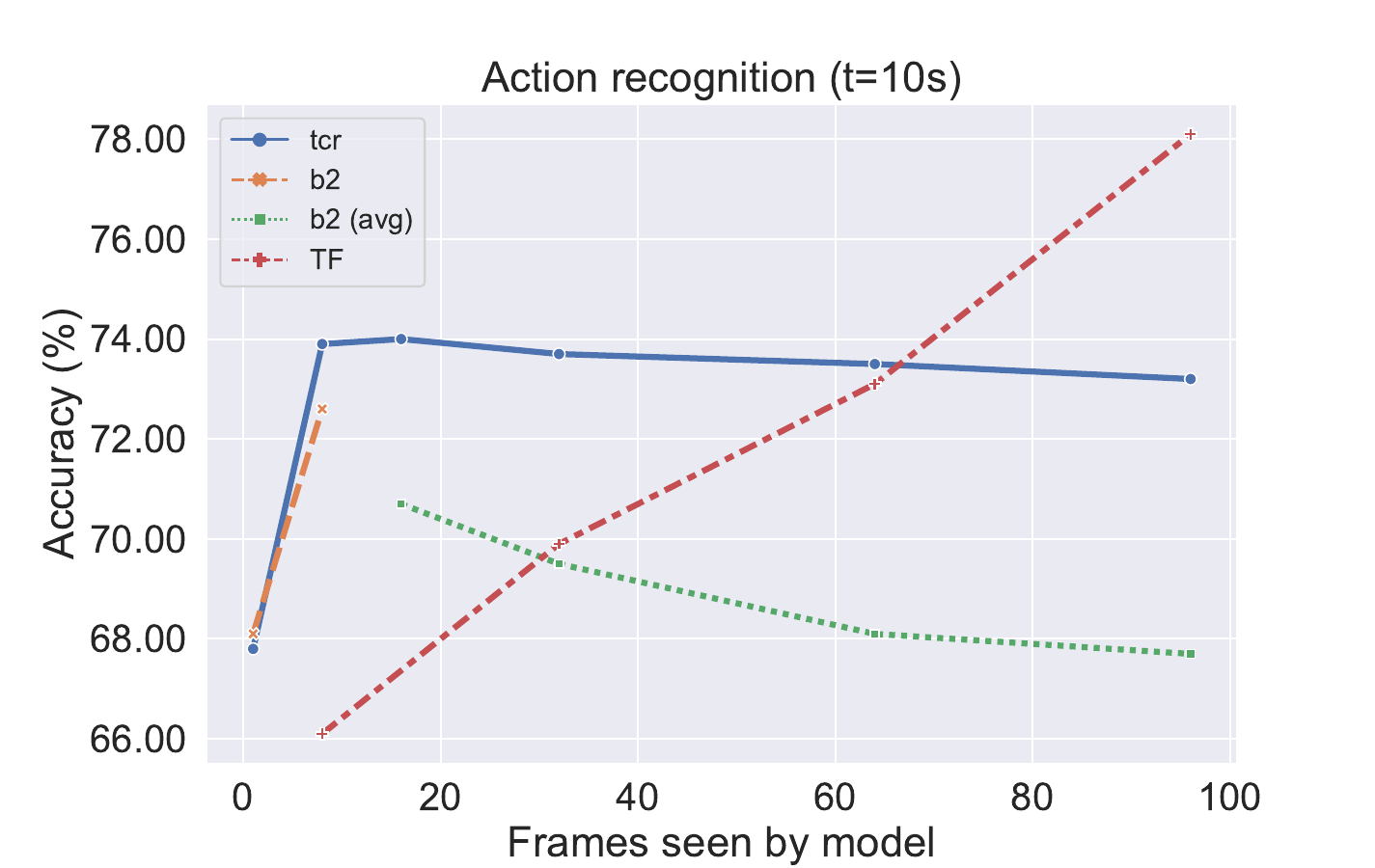}}
    \subcaptionbox{ \label{sfig:cap}}[.3\textwidth]{\includegraphics[height=1in, trim=0 0 0 0,clip]{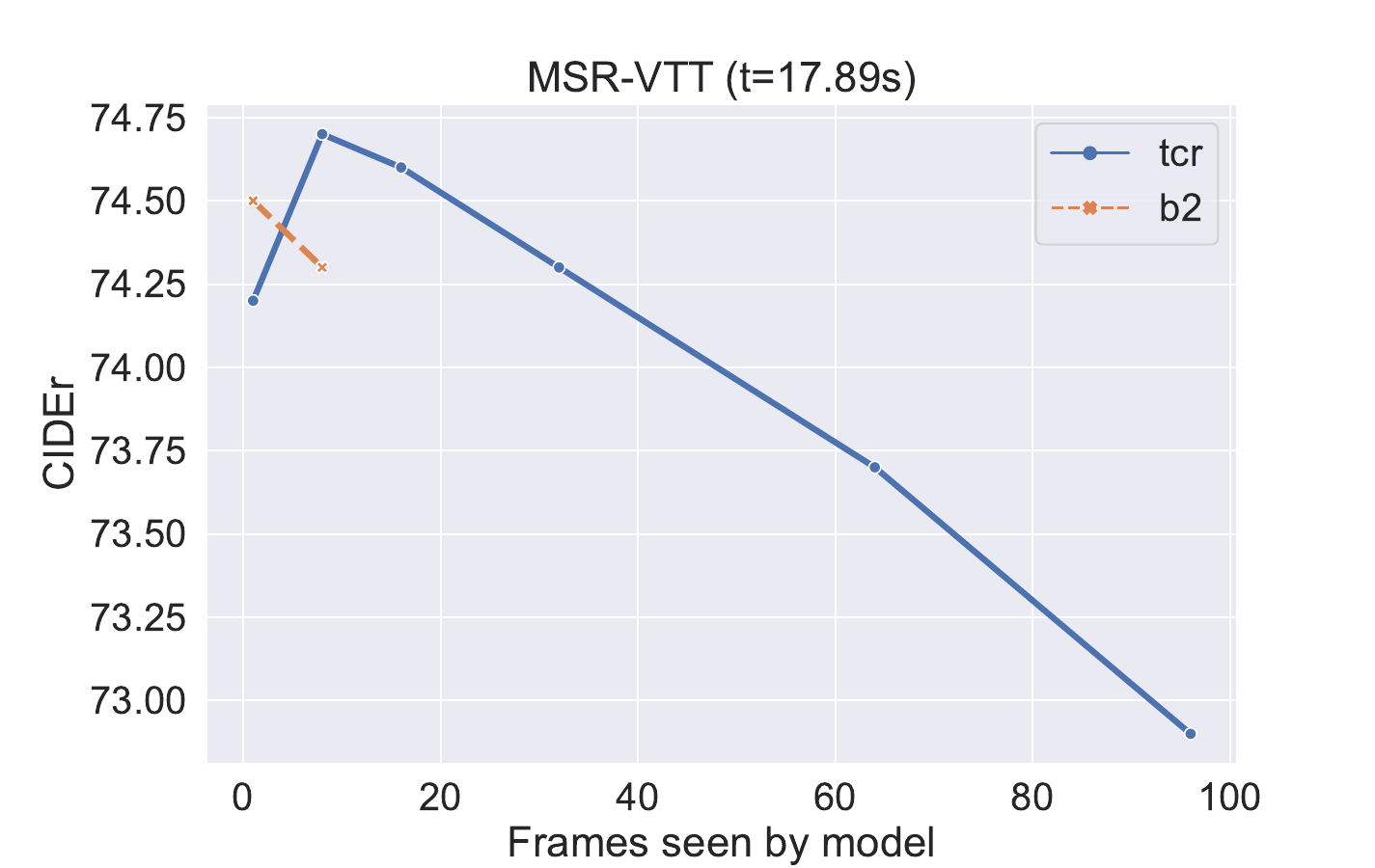}}
    \subcaptionbox{ \label{sfig:lta}}[.3\textwidth]{\includegraphics[height=1in, trim=0 0 0 0,clip]{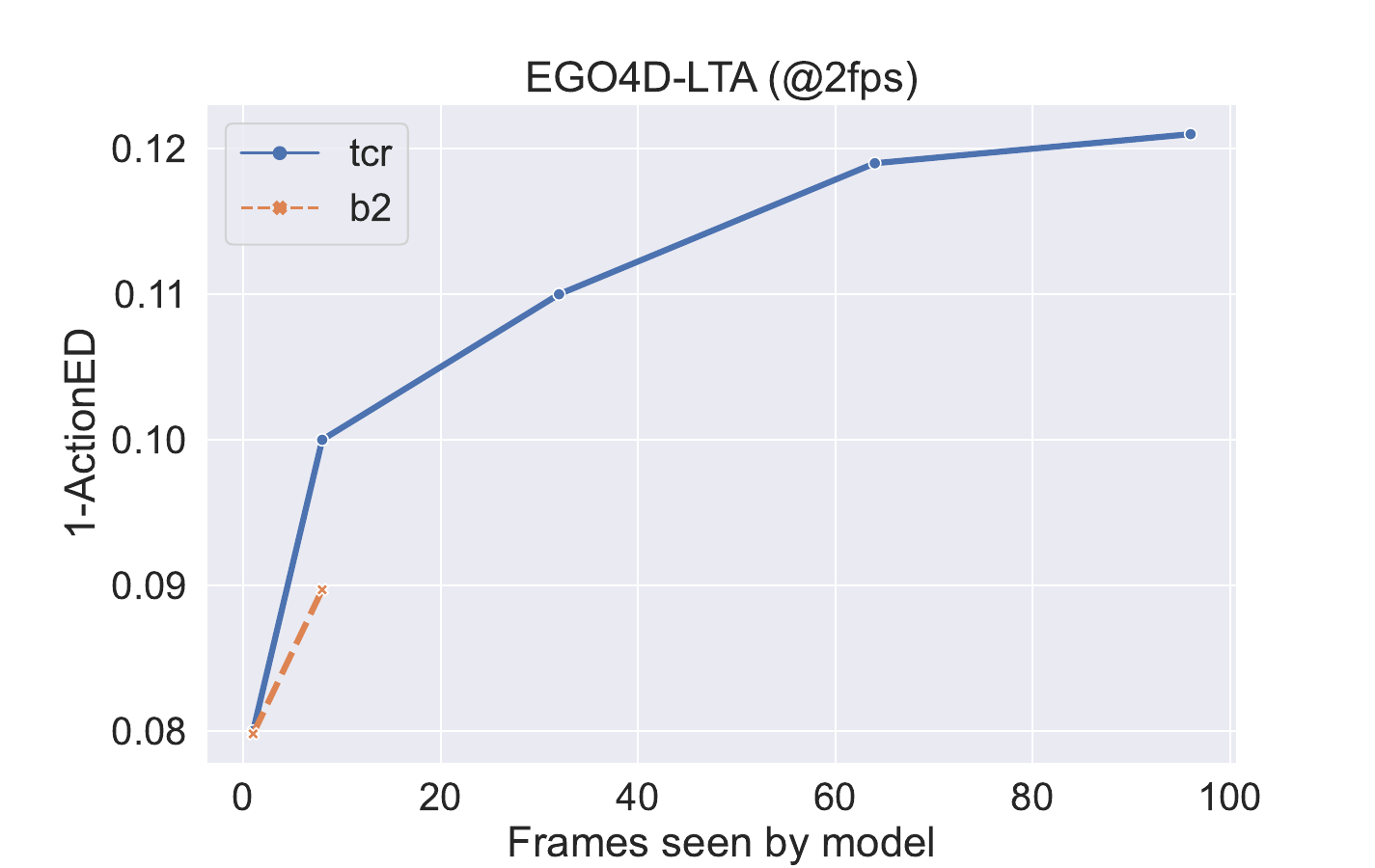}}
    \subcaptionbox{ \label{sfig:count}}[.3\textwidth]{\includegraphics[height=1in, trim=0 0 0 0,clip]{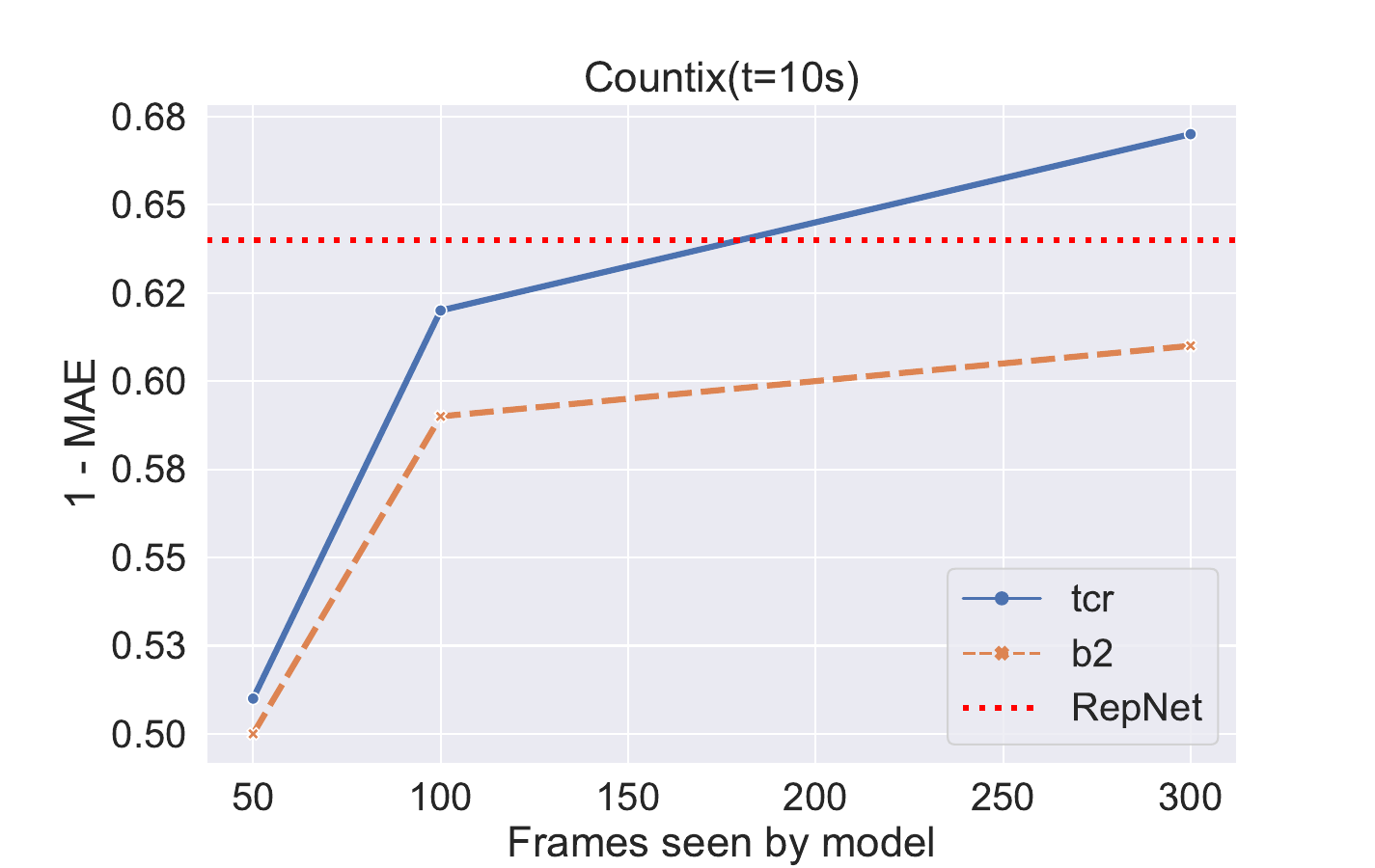}}
\vspace{-5pt}
    \caption{\small Performance vs number of frames utilised by the models on various different tasks. $t$ denotes average length of the video in the dataset. `tcr'=Ours, `iv'=IntenVideo~\cite{wang2022internvideo}, `TF'=TimesFormer~\cite{bertasius2021space}, `b2'=BLIP2~\cite{li_blip-2_2023}, `sevilla'=~\cite{yu2023sevilla}, RepNet=~\cite{dwibedi2020counting}}
\vspace{-15pt}
    \label{fig:temporaltasks}
\end{figure}

Since videos are a highly redundant data-source, one has to ask how many frames, and at what sampling rate, does the model actually \textit{need} to see to achieve good performance. For example, it has been observed that humans solve QA tasks with 8\% higher accuracy when videos are sampled at 25fps as opposed to sampling them at 1fps~\cite{mangalam2023egoschema}.
In this section, we analyse the results on six common video-understanding tasks with respect to the number of frames consumed by the model. We look at the results from a high-level perspective, in order to determine which tasks will require higher number of frames for a model to solve. Overview of the results can be seen in Figure~\ref{fig:temporaltasks}. BLIP2 and TCR models are initialised from the same checkpoint, and then finetuned on the downstream task using the target number of frames as an input. We describe evaluation procedure for each task in more detail in their respective sections.

Intuitively, question-answering is one of the tasks where longer spans and better understanding of temporal dependencies would be of utmost importance. On the NextQA dataset, where questions contain temporal aspects (\ref{sfig:nextqa}), seeing the span of an entire video seems to be crucial for performance. There is a clear peak at about 2fps, and a sharp decline past 1fps. This means that it is important to observe most of the frames, but frame density is not strictly required. 

Curiously, although long video sequences at higher frame-density are required for humans to solve problems in EgoSchema dataset (\ref{sfig:egoschema}), most models' performance actually peak or plateaus at significatly smaller number of frames~\cite{mangalam2023egoschema}. We argue that this is because they have not been trained with long input length, and subsequently fail to capture semantic interdependencies within the video. TCR's performance increases with the number of frames, but plateaus when more sparsity in frame patches is needed to keep memory consumption down (see Table~\ref{tbl:egoschema} for more details). We believe that being able to \textit{see} the span of entire video with more density (i.e.\ at sampling rates greater than 1fps) could further increase performance on this benchmark. 

Human action recognition (\ref{sfig:ar}) and short-video captioning (\ref{sfig:cap}) are commonly used video-understanding benchmarks, however, we found that they do not require many frames to achieve strong performance with a VLM. On action recognition, specialised models such as~\cite{bertasius2021space} scale better with higher frame density, however, which can be attributed to their to inherently learned sampling~\cite{korbar2019scsampler}. This intuition can be corroborated by a slight performance increase when using TCR module with BLIP2.
Future prediction tasks (\ref{sfig:lta}) on egocentric videos often span a long temporal range. We found that increasing the number of frames, hence covering larger video spans, helps in reducing the overall error. This is not unexpected, as action sequences in EGO4D tend to be repetitive, so the longer sequence of actions allows the model to recognise the action pattern from more samples.
Finally, counting (\ref{sfig:count}) is an example of the task where frame density matters, and performance drops when fewer frames are utilised from a fixed length video. It also requires specialised architecture to solve it~\cite{dwibedi2020counting}, and LLMs are traditionally disadvantaged in tasks that require numerical reasoning. For action classification and counting problems, we only use visual and aggregator parts of the VLM as described in the supplementary.

To conclude, although the tasks that require the models to be able to reason over many frames either in span or density are fairly limited, they do exist. Below, we show how a module such as TCR that allows us to `see' more frames in context could be beneficial to overall performance on these tasks. 


\subsection{Evaluation on video question-answering}
\label{sec:nextqa}
We first evaluate our model on question-answering benchmarks, our prior experiment shows a clear benefit when more frames are observed. 
We use the NextQA validation set to compare our model to the current state-of-the-art model which is also based on BLIP2 architecture~\cite{yu2023sevilla}. We follow fine-tuning and evaluation protocol from~\cite{li_blip-2_2023}, with hyperparameters outlined in the supplementary. 


\noindent \textbf{Input design:} Video is subsampled to a target numbers of frames (92 frames at approximately 2fps for the final model), and temporal embeddings are computed accordingly. Conditioning text is formed as ``\texttt{[QA] Question}'' where \texttt{[QA]} is learned special token reserved for VQA tasks. During fine-tuning, the prompt to the LLM is formed following~\cite{yu2023sevilla} as: ``\texttt{[vis. features][question][options] Considering information in frames, select the correct answer}''.


\noindent \textbf{Evaluation procedure: }
During inference, we restrict generation to the answer vocabulary (i.e.\ ``Option A'', ``Option B'', ...), and select the most probable answer.


\begin{table}[b]
\centering
\caption{\small Comparison to SOTA on NextQA dataset. Results are split into non-balanced `causal' (C), `temporal' (T) and `descriptive' (D) questions. The overall accuracy in the last column to the right is balanced across the entire dataset, rather than across the categories. `*' denotes re-implementation by~\cite{sevilla2021only}}
\vspace{-5pt}
\label{table:sota_rest}
\small
\begin{tabular}{@{}lccccc@{}}
\toprule
Model       & train params & accC~$\uparrow$ & accT~$\uparrow$ & accD~$\uparrow$ & acc~$\uparrow$  \\ \midrule
SeViLA~\cite{yu2023sevilla} & 346M   & 73.4  & 68.8  & \textbf{83.5}  & 73.4 \\
HiTeA~\cite{ye2023hitea}  &   /    & 62.4  & 58.3 & 75.6 & 63.1 \\
BLIP2~\cite{li_blip-2_2023}  & 188M   & 64.9  & 59.7  & 77.8  & 63.5 \\
BLIP2*~\cite{li_blip-2_2023, sevilla2021only} & 188M & 72.9 & 65.2 & 80.1 & 70.1 \\
Ours   & 76M    & \textbf{73.5}  & \textbf{69.8}  & 82.2  & \textbf{73.5} \\
\bottomrule
\end{tabular}
\end{table}


\noindent \textbf{Comparison to SOTA:}
Results can be found in Table~\ref{table:sota_rest}. Our model outperforms BLIP2 which demonstrates that the TCR module is successful in selecting the relevant frames, and also indicates the need for temporal modelling of this particular task.
While like us, the SeViLA model is based on BLIP2, they train one model to sample relevant keyframes, and a separate model to solve the task from the sampled keyframes, effectively doubling the number of trainable parameters. In contrast, TCR requires only a single forward pass during training to both sample the features and solve the task. Our model outperforms SeViLA in overall accuracy (setting the new SOTA), hence showing that number of observed frames makes up for lack of trainable parameters.


\subsection{Evaluation on long-form VQA}
\label{sec:egoschema}
EgoSchema is a long-form VQA dataset sampled from EGO4D containing 5000 human curated multiple choice question answer pairs, spanning over 250 hours of real video data. Each question requires the model to select one out of 5 possible answers and is accompanied by a three-minute-long video clip~\cite{mangalam2023egoschema}. Input and evaluation designs are the same as they are for NextQA. 


\noindent\textbf{Comparison to SOTA:}
Results can be seen in Table~\ref{tbl:egoschema}. Our model outperforms both the SOTA models (where inference was done over multiple forward passes and prediction was averaged) and our re-implementation of BLIP2 (with both subsampled frames and iterative inference approach). Similar to~\cite{mangalam2023egoschema}, we observe relative saturation of performance with increasing the number of frames. 

\begin{table}[t]
    \centering
    \caption{\small Comparison to SOTA and human performance on EgoSchema split of EGO4D. $\times$ denotes multiple forward passes were used. * denotes higher proportion of patches was dropped.}
    \label{tbl:egoschema}
    \vspace{-5pt}
    \small
    \begin{tabular}{@{}lcc@{}}
    \toprule
    Method & \begin{tabular}[c]{@{}c@{}}Observed\\ frames\end{tabular} & \begin{tabular}[c]{@{}c@{}}QA\\ acc (\%)~$\uparrow$\end{tabular} \\ \midrule
    InternVideo~\cite{chen2022internvideoego4d}& $8${\scriptsize $\times 11$} & 32.1 \\
    BLIP2~\cite{li_blip-2_2023} & 8 & 27.2 \\
    BLIP2~\cite{li_blip-2_2023} & $8${\scriptsize $\times12$} & 29.9 \\
    TCR (ours) & 92 & 34.2 \\ 
    TCR (ours) & $92${\scriptsize $\times2$} & 34.5 \\
    TCR (ours) & 184* & \textbf{35.1} \\
    \midrule
    Human & 180 & 67.2 \\ \bottomrule
    \end{tabular}
\vspace{-10pt}
\end{table}


\subsection{Evaluation on EGO4D challenges}
\label{sec:ego4d}

\subsubsection{Long-term action anticipation (LTA):} 
The goal of the LTA challenge is to predict a sequence of twenty actions in order of appearance from an input video. The last observed action and action boundaries are given as well. The current state-of-the-art method relies solely on the power of large-language models in order to predict the sequence of future actions~\cite{huang2023palm}. Our model adapts this idea but leverages the ability of TCR to process increasingly longer videos in order to achieve superior results. We compare our model to the SOTA, as well as to fine-tuned BLIP2 using 8 frames as video input. We note that our model outperforms BLIP2 by a significant margin, clearly showing the benefits of being able to observe denser video sequences for this task. Results can be seen in Table~\ref{tbl:sota_ego4dlta}.


\noindent\textbf{Input design:} We construct input for fine-tuning and evaluation in the following fashion: video is subsampled uniformly to a target number of frames (8 for BLIP2 with Q-former, and 96 for BLIP2 with TCR) and temporal embeddings denoting the frame timestamp are added to them. If the ``Before'' video is sampled, we only sample from a video clip before the last observed action, while ``Whole'' means we sample from the entire input video clip. The text prompts are designed as:
\vspace{-3pt}

\begin{center}
\begin{lstlisting}[basicstyle=\small]
         Complete an action sequence, 
         an action is one (verb, noun) pair. 
         A complete sequence consists of 28 actions. 
         Actions: (noun_1, verb_1) (verb_2, ...
\end{lstlisting}
\vspace{-3pt}
\end{center}
\noindent and for the conditioning prompt we use:
\begin{lstlisting}[basicstyle=\small]
[LTA][start_1](noun_1, verb_1),[start_2](noun_2, verb_2) ...
\end{lstlisting}
where \texttt{(noun, verb)} is an action pair, \texttt{[LTA]} is a learned special token, and \texttt{[start k]} is a tokenised start time of $k$-th action.


\noindent\textbf{Evaluation procedure:} Our evaluation procedure follows closely those of~\cite{huang2023palm}. The model is fine-tuned to output comma-separated action pairs following the prompt formatting. During the evaluation, we softmax predictions over the reduced vocabulary of the label space for the LTA task. If both nouns and verbs fall into their respective label space, we append them to our prediction. For predictions with less than 20 action pairs, we pad it with the last action. Models denoted with (*) are sampled in an iterative fashion. 


\noindent\textbf{Comparison to SOTA:} Table~\ref{tbl:sota_ego4dlta} shows the comparison to the state-of-the-art on long-term action prediction. Note that SOTA~\cite{huang2023palm} uses only language model to predict the future actions. We demonstrate that being able to perceive frames after the indicated timestep (which are given) helps for future action prediction, but we outperform sota even without seeing them. Finally, we find that iterative evaluation (i.e. asking the model to predict action by action, as opposed to the whole set of 20 actions, increases performance even further.  

\begin{table}[b]
\centering
\caption{\small Comparison of various models on the validation set of {\em EGO4D LTA challenge}. Edit distance is reported and the lower the score the better. The ``Video'' column indicates whether the whole video was observed (given) or just the video clip before the last action. Models denoted with `*' are sampled iteratively.}
\label{tbl:sota_ego4dlta}
\small
\begin{tabular}{@{}lcccc@{}}
\toprule
Method & Video & VerbED~$\downarrow$ & NounED~$\downarrow$ & ActionED~$\downarrow$ \\ \midrule
PALM*~\cite{huang2023palm} & {\footnotesize No} & 0.7165 & 0.6767 & 0.8934 \\
BLIP2~\cite{li_blip-2_2023} & {\footnotesize Before} & 0.7512 & 0.6873 & 0.9103 \\
BLIP2~\cite{li_blip-2_2023} & {\footnotesize Whole} & 0.7500 & 0.6799 & 0.9086 \\
Ours & {\footnotesize Before} & 0.7009 & 0.6472 & 0.8792 \\
Ours & {\footnotesize Whole} & 0.6763 & 0.6180 & 0.8522 \\
Ours* & {\footnotesize Whole} & \textbf{0.6585} & \textbf{0.6171} & \textbf{0.8482} \\ 

\bottomrule
\end{tabular}
\end{table}


\subsubsection{Moment queries (MQ):} 
The MQ task is similar to temporal action localisation or moment retrieval tasks. Given a textual description of an action, the goal is to localise all possible instances of it in the given video clip. Results can be seen in the Table~\ref{tbl:sota_ego4dmq}.


\noindent\textbf{Input design:} Video is subsampled uniformly to the target number of frames, and temporal embeddings denoting the frame timestamps are added to it. The conditioning prompt is formed as ``\texttt{[TRG] action query string}'' where \texttt{[TRG]} indicates the special token for temporal grounding and \texttt{action query string} denotes the name of the action label parsed as a string. The language model is prompted by the following string 
\begin{lstlisting}[basicstyle=\small]
         Return a sequence of frame timestamps 
         where <action name> is happening. The
         timestamps range from 1 to 1000.
\end{lstlisting}


\noindent\textbf{Evaluation procedure:} We softmax the model predictions from a reduced vocabulary of integers from 1 to 1000 (temporal coordinates are quantised similarly to~\cite{chen2021pix2seq}) and aggregate them.


\noindent\textbf{Comparison to SOTA:} Results in Table~\ref{tbl:sota_ego4dmq} show that despite the disadvantage of solving a discriminative task in a generative way, our model still performs admirably (-2.4 MAP) when compared to the state-of-the-art. In the supplementary material, we present an additional evaluation procedure (using direct classification) which can yield even better performance (+1.25 MAP).
\begin{table}[t]
    \centering
    \caption{\small Comparison to the state of the art on the validation set of {\em Ego4D Moment Query Challenge}.}
    \label{tbl:sota_ego4dmq}
    \small
    \begin{tabular}{@{}lcc@{}}
    \toprule
    Method & Avg. mAP~$\uparrow$ & R@1, tIoU=0.5~$\uparrow$ \\ \midrule
    Intern Video~\cite{chen2022internvideoego4d} & 23.59 & 41.13 \\
    ASL~\cite{shao2023asl} &\textbf{ 27.85} & \textbf{46.98} \\
    Ours (96f) & 24.51 & 42.99 \\
    Ours (192f) & 25.45 & 43.72 \\ 
    \bottomrule
    \end{tabular}
\end{table}


\subsection{Model design decisions} 
\label{sec:ablation}

\begin{table}[b]
    \centering
\caption{\small Ablation studies on validation set of the NextQA dataset. Note that the ablations were done on a short training schedule.}
\label{tbl:ablation}
    \begin{subtable}{0.25\textwidth}
    \small
    \centering
    \begin{tabular}{@{}lc@{}}
    \toprule
    cond. & acc~$\uparrow$ \\ \midrule
    yes & \textbf{64.9} \\
    none & 61.1 \\
    corrupt & 55.3 \\ \bottomrule
    \end{tabular}
    \caption{\scriptsize Different conditioning prompts on \textit{temporal-question set only}.}
    \label{table:abl_cond}
\end{subtable} \hspace{10pt}
\begin{subtable}{0.25\textwidth}
    \small
    \centering
    \begin{tabular}{@{}lc@{}}
    \toprule
    \#frms & acc~$\uparrow$ \\ \midrule
    32 & 64.4 \\
    92 & \textbf{66.2} \\
    124 & 65.9 \\ 
    \bottomrule
    \end{tabular}
    \caption{\scriptsize Impact of number of frames on model performance.}
    \label{table:abl_numframes}
\end{subtable}\hspace{10pt}
\begin{subtable}{0.33\textwidth}
    \centering
    \small
    \begin{tabular}{@{}ll@{}}
    \toprule
    \#queries & acc~$\uparrow$ \\ \midrule
    32 & 62.7 \\
    64 & 65.8 \\
    128 & \textbf{66.2} \\
    256 & 64.3 \\ \bottomrule
    \end{tabular}
    \caption{\scriptsize Impact of the total number of queries on model performance.}
    \label{table:numqueries}
    \end{subtable}
\end{table}

In the following section we investigate our model choices and seek to explain their impact on the performance of our model. All experiments were done on the validation set of NextQA dataset and results can be seen in Table~\ref{tbl:ablation}. Note that we fine-tune the model on a shorter training schedule which yields lower results, but allows for a quicker turnaround. We keep the same fine-tuning parameters for all ablation studies.


\noindent\textbf{Does text conditioning impact the results?} We investigate the performance of our model in three different scenarios: (1) when the conditioning prompt is unchanged in the evaluation setting, (2) we completely remove the conditioning prompt, and (3) we modify the temporal word (`before' to `after' and vice-versa) in a hope to confuse the model. The results can be seen in Table~\ref{table:abl_cond}. Conditioning indeed allows the TCR module to extract more relevant features (+3.8). Furthermore, adversarial conditioning greatly impacts the performance of the model (-7.6). 


\noindent\textbf{Do we need special tokens for conditioning? }If the model is fine-tuned for a specific task without the special tokens, it still performs reasonably well ($73.5\%$ vs $72.7\%$ acc on NextQA with and without special tokens respectively).


\noindent\textbf{Does the number of frames matter?} The input video-sequence length is important to the model performance. In Table~\ref{table:abl_numframes} we show the performance dependence on the input sequence length. Note that videos are on average 44s long, thus 124 frames equates to sampling at a rate of 2.5fps.


\noindent\textbf{How many queries should the LLM see?}  While there is a benefit of perceiving a longer length of a video input-sequence, it has been observed that including more visual tokens as input to the LLM does not lead to a better performance~\cite{yu2023sevilla}. Therefore in Table~\ref{table:numqueries} we investigate how many queries the LLM should observe. Reducing the number of queries to a total of 128 (equivalent to four frames according to~\cite{li_blip-2_2023}) achieves optimal performance.

\section{Related work}
\label{sec:relwork}

Our work spans many fields of video-understanding and we outline the most relevant related work below.


\noindent \textbf{Video-sampling techniques: }
Sampling relevant frames from videos has long been a challenge in video understanding due to the highly redundant nature of video data. These methods either use a pre-processing module~\cite{chen2011dynamic, yeung2016end, korbar2019scsampler, gowda2021smart, wang2021adaptive, zhi2021mgsampler, buch2022revisiting} to guide their model through multi-modal attention-like mechanisms~\cite{gao2020listen, panda2021adamml}, or employ recursive reinforcement learning techniques~\cite{wu2019liteeval} to select relevant parts of the videos. 
In temporal action detection, models are tasked with precisely locating action boundaries. Most commonly used datasets~\cite{zhao2019hacs, wang2014thumos, caba2015activitynet} are dominated by custom solutions or transformer architectures~\cite{shao2023asl, zhang2022actionformer} built upon strong features~\cite{tran2018closer, wang2022internvideo, carreira2017quo, tong2022videomae, wang2023videomae}. 


\noindent\textbf{Egocentric videos understanding:}
Extreme length and temporally sensitive nature of the tasks introduced in EGO4D required researchers to think about the problem of video-length on a different scale~\cite{Ego4D2022CVPR, mavroudi2023learning}. This has already yielded creative approaches to solve various challenges in the egocentric space~\cite{tan2023multiscale, jiang2023single, chen2022internvideoego4d}. Also new and exciting benchmarks have been developed: for example the recent EgoSchema dataset~\cite{mangalam2023egoschema}, a manually annotated subset of EGO4D where each QA pair corresponds to a 3 minute-long video. A work particularly relevant to ours is SpotEM~\cite{ramakrishnan2023spotem}, a lightweight sampling mechanism that makes use of low dimensional features in order to select video segments important for solving natural-language query challenge . Though impressive in performance, their method is limited to the type of embeddings used for sampling, and is thus less general than our approach.

\noindent \textbf{Video-language models and feature resampling: }
VLMs have revolutionised the field of computer vision -- the scale of the models and data they were trained on increased exponentially in a short period of time~\cite{alayrac_flamingo_2022, zellers_merlot_2021, jia_scaling_2021, li_blip-2_2023, radford2021clip}, some even being jointly optimised for images and video~\cite{alayrac_flamingo_2022, kuo_mammut_2023, li_blip-2_2023}. The length of the videos these models can process often varies -- \cite{alayrac_flamingo_2022} can process up to 8 frames, \cite{kuo_mammut_2023} can process longer tubelets (at reduced receptive fields). None of these models can process videos over 16 frames at full resolution outright. 

\noindent \textbf{Concurrent works on VLMs for video: }
\
Extending capabilities of VLMs is a fast-paced area of research, and many works have appeared without being published. \cite{li2023llamavid} conducted an orthogonal study, exploring how the embedding quality can reduce the amount of visual information necessary for large LLMs. We on the other hand introduce a bottleneck module to increase the amount of data processed without increasing complexity. \cite{maaz2023videochatgpt} focuses on interactive aspects by improving the LLM pipeline. We keep the pipeline fixed to seek improvements from the data. \cite{zhang2023videollama} increases the amount of information by introducing additional modalities which would be an interesting next step for our work as well.
Similar to us,~\cite{song2023moviechat} aims to increase video length in a BLIP2 model, but they do so via a memory bank. Combining a memory-augmented approach with reasoning over longer sequences would be a promising future work. Techniques like FlashAttention~\cite{dao2022flashattention} or RingAttention~\cite{liu2023ring} also allow the context window of a VLM to handle long sequences of frames but at the cost of significant growth in inference speed. These techniques are complementary to our proposal and could be integrated in the TCR to support even longer videos in the future.


\section{Conclusion}
We present a parameter-efficient, text-conditioned module and training method for bridging video-to-text gap that can be applied to a large number of frames in videos. Even though our model is entirely based on BLIP2~\cite{li_blip-2_2023} architecture, introducing it in other VLMs would be straightforward. We believe that models capable of perceiving long video sequences such as TCR will open up a promising new direction in research. 

%% file: _supp.tex
\section*{Contents}
\noindent \begin{enumerate}
    \item[] \ref{sec:model}~Further model details
    \begin{enumerate}
        \item[] \ref{sec:11}~Implementation details
        \item[] \ref{sec:12}~Time tokenisation
    \end{enumerate}
    \item[] \ref{sec:traindeedts}~Further training details
    \begin{enumerate}
        \item[] \ref{sec:21}~Pre-training
        \item[] \ref{sec:22}~Fine-tuning procedure
    \end{enumerate}
    \item[] \ref{sec:standalone}~Study of standalone TCR architecture
    \begin{enumerate}
        \item[] \ref{subsec:poc}~Study of TCR's sampling capabilities
        \item[] \ref{subsec:tcrnollm}~TCR without LLM
    \end{enumerate}
\end{enumerate}

\section{Further model details}
\label{sec:model}

\subsubsection{}{Implementation details: }
\label{sec:11}
The model is implemented in FLAX~\cite{flax2020github}, based on the scenic framework~\cite{dehghani2021scenic}. We use BLIP2 ViT-g Flan$T5_{xl}$ as a starting point and keep the vision and text models frozen. The number of \textit{trainable} parameters is about 1\% of the total memory footprint. JAX allow us to avoid storing gradients for non-trainable parameters, thus freeing up additional memory. Without training any part of the VLM, the model can process up to 124 frames during training time per TPU.

\subsubsection{Time tokenisation: }
\label{sec:12}
In order to make the model time aware, we add time tokens to text and video-frames.
We tokenise the time in half-second increments following an analogue procedure from~\cite{chen2021pix2seq} with $n_{\text{bins}}=2048$. This means we can tokenise up to 17 minutes of video with a precision of half a second. We use this time format to express the temporal localisation of a segment in a longer video if that information is required (for example in moment retrieval or when describing a segment of the video). For each frame, we pass its timestep, tokenised as above, through a single-layer MLP in order to obtain learnable temporal embeddings we add to every frame.

\section{Further training details}
\label{sec:traindeedts}

Section~2.2 of the main paper give the purpose of each training stage, and what is done, while Table~1 in the main paper investigates the impact of each stage on the downstream tasks. In this section, we first detail the pre-training stage and illustrate different tasks, then we give the  fine-tuning details, and finally investigate 0-shot performance of multi-task trained models. All models are trained using the AdamW optimizer with global norm clipping of 1. Training hyperparameters are given in Table~\ref{tab:hyperparams}.

\subsection{Pre-training}
\label{sec:21}
\begin{figure}[t]
    \centering
    \includegraphics[width=.6\textwidth]{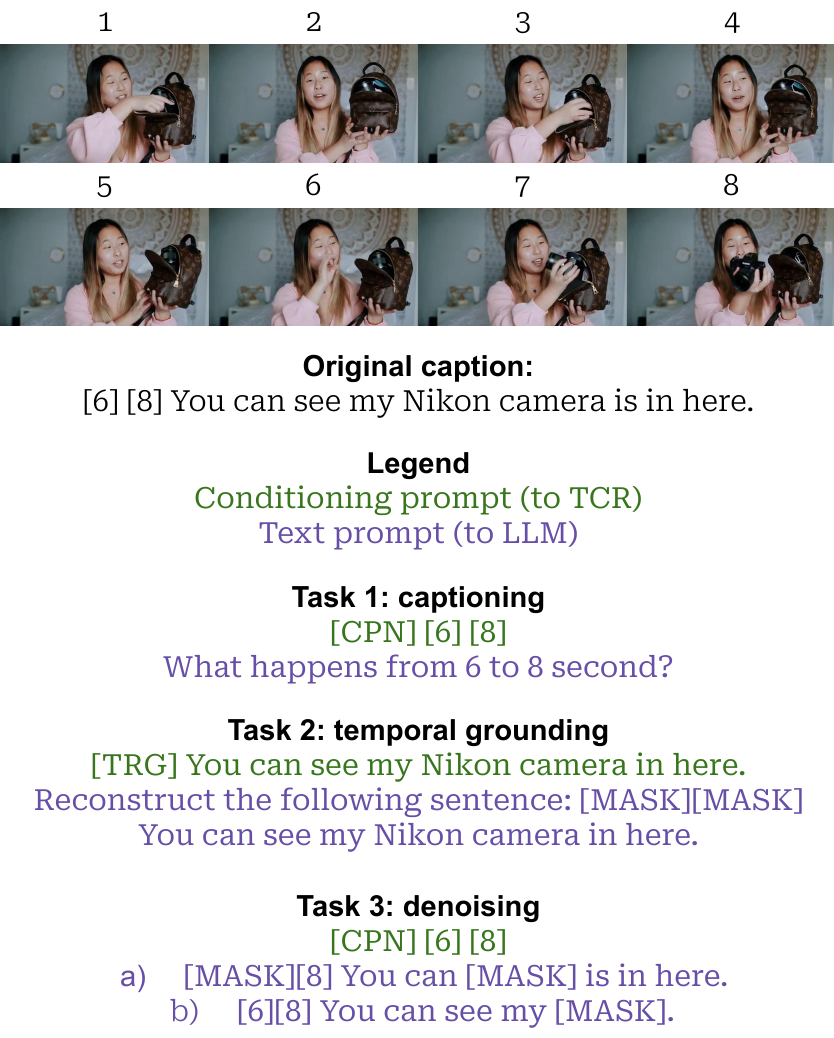}
    \caption{Pre-training task examples with a condition sequence (TCR input) and the context sequence (LLM input). [CPN] (captioning), [TRG] (temporal grounding) are special tokens, and [6] is a sample of a tokenised timestamp. We use [MASK] (masking) token to form LLM prompts where applicable as they are integral part of T5's training~\cite{raffel2020exploring}. The model is tasked with predicting a caption ``You can see my Nikon camera is in here'' which is happening between 6th and 8th second of a video.}
    \label{tbl:tasks}
\end{figure}

After initialisation, we pre-train the model using the three outlined tasks (captioning, temporal grounding, denoising). Specifically, we define a dataset with each task, assign it a weight (1.0, 0.5, 0.5 respectively), and then, following procedure in~\cite{alayrac_flamingo_2022}, we accumulate gradients across each task with loss being weighted per-dataset. Loss weights were defined empirically and kept fixed throughout. An illustration of dataset tasks can be seen in Figure~\ref{tbl:tasks}.

\subsection{Fine-tuning procedure}
\label{sec:22}

Due to a diversity in our downstream tasks, both in terms of the data type (e.g.\ instructional videos, vs egocentric videos) and tasks for which we train special tokens, devising a model that can follow specific dataset conventions or match a particular answer format from noisy pre-training data is a challenging task. We found that fine-tuning the model alleviates these challenges and helps our model's performance. If heplful, we introduce a new special token for the task, and proceed to fine tune the model on the downstream task specifically.

\begin{table*}[t]
\small
\centering
\begin{tabular}{@{}llcccc@{}}
\toprule
Stage & Dataset & Batch size & LR & Epochs & Warmup steps \\ \midrule
\multirow{3}{*}{Pre-training} & YTT - captioning & \multirow{3}{*}{256} & \multirow{3}{*}{1e-4} & \multirow{3}{*}{10} & \multirow{3}{*}{1k} \\
 & YTT - temporal grounding &  &  &  &  \\
 & YTT - denoising &  &  &  &  \\
 \midrule
\multirow{6}{*}{Fine-tuning} & MSR-VTT & 128 & 1e-5 & 20 & 2k \\
 & NextQA & 128 & 1e-5 & 40 & 2k \\
 & NextQA - short & 64 & 5e-5 & 10 & 1k \\
 & EgoSchema & 64 & 1e-5 & 20 & 1k \\
 & EgoLTA & 64 & 1e-5 & 20 & 1k \\
 & Ego4D MQ & 128 & 1e-5 & 10 & 500 \\ \bottomrule 
\end{tabular}
\caption{Pre-training and fine-tuning hyper-paramerters}
\label{tab:hyperparams}
\end{table*}

\noindent\textbf{EgoSchema fine-tuning: }
EgoSchema~\cite{mangalam2023egoschema} dataset does not come with a dedicated training set. We use EGO4D narrations to fine-tune the model as follows. We first identify video IDs from EGO4D that are also present in EgoSchema benchmark and remove them. We also remove all videos where narrations are tagged with an `\#unsure' tag. We then fine-tune the model with a template ``What does the {templ} do from {start} to {end}?'' if narration is tagged with `\#C' or `\#O', or ``What happens from {start} to {end}'' if the narration is tagged with `\#summary'. Fine-tuning hyper-parameters can be found in Table~\ref{tab:hyperparams}.

\subsubsection{Zero-shot performance and multi-task fine-tuning.}
Models such as Flamingo~\cite{alayrac_flamingo_2022} can be `prompted' to solve tasks in a low-shot setting, thus making fine-tuning unnecessary. Flamingo, however, is trained on much wider variety of datasets (mostly from an image domain, where large datasets with less noisy supervision are available). Our model lacks this pre-training and, additionally, faces a significant challenge in this setting, as the number of output queries from TCR (128) is significantly higher than that of Perciever Resampler (8) resulting in longer sequences.
However, having a single general model capable of solving multiple tasks is an appealing proposition. Therefore, we present the results with and without fine-tuning in $0$-shot and $k$-shot setting in Table~\ref{table:finetuning}. First two rows use a model initialised and pre-trained without any modifications. Second two rows show the performance of a {\em single} model (one set of weights) that is trained on all downstream datasets jointly with equal weighting. In other words, after pre-training, we fine-tune the model jointly on \textit{all} downstream datasets for 20 epochs, accumulating the gradients over all datasets with equal weight. Finally, in the last two rows, we present the results of four fully fine-tuned models. As observed in models utilising large instruction-tuned language models, our model performs better in a few-shot setting~\cite{alayrac_flamingo_2022}. As expected the improvement in fully fine-tuned models is minor since models are already specialised for the specific task. We want to highlight how our model with a single set of weights can perform well across different tasks simply with the addition of a few prompts for in context learning. In case even stronger performance are needed, we can further improve by specialising the model for the specific task/dataset, and we report these numbers in the main paper.

\begin{table}[t]
\small
\centering
\begin{tabular}{@{}lccccc@{}}
\toprule
FT &
  k-shot &
  \begin{tabular}[c]{@{}c@{}}NextQA\\ (Acc)\end{tabular} &
  \begin{tabular}[c]{@{}c@{}}Ego4D-LTA\\ (ActionED)\end{tabular} &
  \begin{tabular}[c]{@{}c@{}}Ego4D-MQ\\ (avg mAP)\end{tabular} &
  \begin{tabular}[c]{@{}c@{}}EgoSchema\\ (acc)\end{tabular} \\ \midrule
\multirow{2}{*}{None}      & 0 & 34.2 & 0.9354 & 19.88 & 14.5 \\
                           & 2 & 37.3 & /      & 21.75 & 19.0 \\
\multirow{2}{*}{Multitask} & 0 & 61.7 & 0.9199 & 21.74 & 28.7 \\
                           & 2 & 64.1 & /      & 23.01 & 30.9 \\
\multirow{2}{*}{Per task}  & 0 & 73.5 & 0.8782 & 24.51 & 34.2 \\
                           & 2 & 73.5 & /      & 24.83 & 34.1 \\ \bottomrule
\end{tabular}
\caption{Comparison of pre-trained vs fine-tuned model on the downstream datasets in 0- and few-shot setting. Numbers reported in the main comparison are fine-tuned per-task and evaluated in a 0-shot setting for consistency. Note that for Ego4D-LTA task, we couldn't fit the entire action sequence of a multi-shot example due to the maximum sequence length of our LLM.}
\label{table:finetuning}
\end{table}

\section{Study of standalone TCR architecture}
\label{sec:standalone}

In order to further validate our architecture choice, below we conduct a series of experiments with various uses of TCR without the LLM. In section~\ref{subsec:poc}, we run experiments with TCR simply acting as a ``head'' for a visual model. In this case, the visual backbone is fixed and TCR is trained from scratch on a target task without special tokens. In section~\ref{subsec:tcrnollm}, we initialise and pre-train TCR as described in Section~2.2 of the main paper, and instead of connecting it to an LLM, we fine-tune TCR with a two-layer MLP on top of it. 

\subsection{Study of TCR's sampling capabilities}
\label{subsec:poc}

Transformer decoder modules have been shown to perform well on action-detection tasks~\cite{Han22a}, and several works showed they can cross-attend image information with semantic information~\cite{kamath2021mdetr, korbar2022end}. In this section, we seek to validate our architecture design on a set of simplified tasks. Specifically  we show that \textit{a)} TCR can be trained independently to sample relevant features for action classification,  \textit{b)} `select' the frames corresponding to the query action within a couple of videos, and finally \textit{c)} perform semantic tasks such as counting on actions. We conduct these experiments on various splits of the Kinetics dataset~\cite{kay2017kinetics, dwibedi2020counting}.

\noindent\textbf{Sampling for action recognition: }First, we train the model for action recognition. Most models average predictions over multiple uniformly-sampled frames or short clips in order to obtain a video-level prediction~\cite{tran2018closer}. Dedicated samplers such as~\cite{korbar2019scsampler,zhi2021mgsampler} have been trained to specifically sample relevant frames or clips. The goal of this experiment is to see whether TCR can sample features in such a way to improve overall video-level accuracy. In other words, given a 300 frames input video, can the model sample 32 frames in such a way that classification performance is better than other naive benchmarks. To do so, we extract frame-level features from the frozen ViT-g model, add sinusoidal positional embeddings, and pass them to TCR as a key-value pair to the cross attention layers. We ``prompt'' the model with 32 learnable queries, each of which is processed through a 400-way classifier whose predictions are then averaged to obtain the final prediction. This model is trained for 10 epochs on Kinetics-400 dataset and compared to: linear classification of features from 32 random frames, self-attention pooling where the classification layer is done on a learned \texttt{CLS} token, and on two top video clips sampled by SCSampler~\cite{korbar2019scsampler} (32 frames). TCR is able to outperform naive sampling methods and bests standard attention-pooling approach, which means it can learn about relative importance of input features to the text prompt. It, however, cannot best models which were trained with explicit saliency supervision such as SCSampler. This experiment indicates that incorporation of explicit saliency supervision such as in~\cite{korbar2019scsampler, zhi2021mgsampler, yu2023sevilla} is a promising direction of further research. Results can be seen in Table~\ref{tbl:kineticsac}.

\begin{table}[t]
\small
\centering
    \begin{tabular}{@{}lc@{}}
    \toprule
    Sampling method &  Accuracy $\uparrow$\\ \midrule
    random & 74.3 \\
    Attention & 75.2 (+0.9) \\
    TCR (ours) & 75.7 (+1.4) \\
    SCSampler~\cite{korbar2019scsampler} & 77.8 (+3.4) \\ \bottomrule
    \end{tabular}
    \caption{Standalone TCR (without using a LLM): Linear classification of 32 frames sampled from the Kinetics400~\cite{kay2017kinetics} val set using various sampling methods.}
    \label{tbl:kineticsac}
\end{table}

\noindent\textbf{Text-conditioned sampling: } Next, we want to verify whether the model can ``select'' correct frames from a video given an action class prompt. To this end we design an artificial experiment by stitching 128 frames of up to 8 kinetics videos, and the task is to classify frames containing the prompt action as positive examples. Only a single video of a query class can be present, it's location within the video is randomised, and the number of frames sampled from each video is random (from 1 to 64). We add position encoding (from 0 to 127) to each frame, and feed them to TCR. For query input, we use the action prompt followed by 32 learnable queries. After TCR, we pass each query through 128 one-vs-all binary classifiers, one for each of the input frames. The classifiers are trained to predict the presence or absence of the action at the corresponding input frame. At inference time, if the confidence score for classifier $j$ is higher than $0.5$ we add to the predictions of query $k$ the frame $j$ (note that each query can predict more than one frame). Finally we consider the set of all predictions across all the queries and compare it to the ground truth frame ids. If the frame ids of the frames corresponding to the action prompt are contained in the predicted set we count that as a positive paring and the rest as negative. We compare this TCR approach to simply training the per-frame 400-way action classifier on top of ViT, and count the label as a positive if the target class is predicted. We were able to achieve precision of $0.75$ and recall of $0.68$, which is higher than simply classifying action label for each frame of the video ($0.70$, $0.44$). 

\noindent\textbf{Text-conditioned semantic processing: }
Finally, with minor addition, we show that TCR architecture can be used as a counting mechanism as well due to its potential to ``look'' and reason over every single frame. We consider the settings of~\cite{dwibedi2020counting}, feed the video at full temporal resolution, and form the query the same as in previous paragraph. We then concatenate the queries and pass them through a TransRAC Period Predictor~\cite{hu2022transrac} to obtain the final count. The TCR module and TransRAC predictor are trained on syntetic data as descibed in~\cite{dwibedi2020counting} for 24 epochs. Results outperform the SOTA benchmark, as can be seen in Table~\ref{tbl:countix}. 

\begin{table}[t]
\small
\centering
    \begin{tabular}{@{}lcc@{}}
    \toprule
    Method & MAE $\downarrow$ & OBO $\downarrow$ \\
    \midrule
    \cite{dwibedi2020counting} & 0.36 & 0.30 \\
    TCR & 0.33 & 0.28 \\
    \bottomrule
    \end{tabular}
    \caption{Standalone TCR (without using the LLM): Counting  repetitions in videos on the Countix~\cite{dwibedi2020counting} dataset.}
\label{tbl:countix}

\end{table}

\subsection{TCR without LLM}
\label{subsec:tcrnollm}

In this section, we show that the pre-trained TCR module can be easily adapted to solve discriminative tasks without the need for an LLM. The difference to the previous section is the fact that TCR is initialised and pre-trained using the procedure outlined in Section~2.2 of the main paper. For this set of experiments, we keep the pre-trained visual encoder and TCR module frozen and train simple 2-layer MLP adapters on top of them.
For the EGO4D moment query task, we adapt the alignment module from~\citet{Han22a} which trains two linear layers on top of TCR output with set-prediction loss consisting of two elements: one predicting whether the sequences correspond to a given action and one regressing the temporal span of the query. For the natural language query task, we adopt the same training procedure as~\citet{yang2022tubedetr} for spatiotemporal localisation on top of TCR output. Results can be seen in Table~\ref{tbl:tcr_disc}. It is notable that, without additional pre-training of the TCR module, the features it outputs can be generalised well to tasks that benefit from video sampling. Our general pre-trained architecture comes close to the specialised solution~\cite{shao2023asl} to the Moments Query challenge ($-1.14$ Avg.MAP), giving us future direction for improving it for tasks of video retrieval. Similarly on the Natural Language Query challenge, with only a small adapter trained for spatiotemporal localisation our model challenges the state-of-the-art~\cite{ramakrishnan2023spotem} ($-0.14$ MR@1) which has a sampler and detector head trained for this task specifically. We believe that optimising VLMs for regressive tasks requiring high spatiotemporal resolution will be an important future research direction.

\begin{table}[t]
\centering
\small
\begin{tabular}{@{}lcc|cc@{}}
\toprule
\multirow{2}{*}{Method} & \multicolumn{2}{c|}{MQ} & \multicolumn{2}{c}{NLQ} \\ \cmidrule(l){2-5} 
 & AVG MAP & R@1 & MR@1 & MR@5 \\ \midrule
SpotEM~\cite{ramakrishnan2023spotem} & / & / & \textbf{11.56} & 19.90 \\
ASL~\cite{shao2023asl} & \textbf{27.85} & \textbf{46.98} & / & / \\ 
TCR & 26.71 & 44.96 & 11.42 & \textbf{19.95} \\
TCR w/ LLM & 25.45 & 43.72 & 10.12 & 18.99 \\ 
\bottomrule
\end{tabular}
\caption{Performance of TCR as a standalone module on discriminative downstream tasks. LLMs tend to struggle with regression tasks~\cite{chen2021pix2seq} and while we present these numbers in the main paper due to their flexibility to solve a wide variety of task, we also show that TCR can provide better performance for these tasks using ad-hoc regression adapters and significantly narrowing the gap with state of the art methods.}
\label{tbl:tcr_disc}
\end{table}